# Regularization-based Continual Learning for Fault Prediction in Lithium-Ion Batteries

Benjamin Maschler [a,*], Sophia Tatiyosyan [a], Michael Weyrich [a]

[a] *University of Stuttgart, Institute of Industrial Automation and Software Engineering, Pfaffenwaldring 47, 70569 Stuttgart, Germany*

\* Corresponding author. Tel.: +49 711 685 67295; Fax: +49 711 685 67302. *E-mail address:* benjamin.maschler@ias.uni-stuttgart.de

**Abstract**

In recent years, the use of lithium-ion batteries has greatly expanded into products from many industrial sectors, e.g. cars, power tools or medical devices. An early prediction and robust understanding of battery faults could therefore greatly increase product quality in those fields. While current approaches for data-driven fault prediction provide good results on the exact processes they were trained on, they often lack the ability to flexibly adapt to changes, e.g. in operational or environmental parameters. Continual learning promises such flexibility, allowing for an automatic adaption of previously learnt knowledge to new tasks. Therefore, this article discusses different continual learning approaches from the group of regularization strategies, which are implemented, evaluated and compared based on a real battery wear dataset. Online elastic weight consolidation delivers the best results, but, as with all examined approaches, its performance appears to be strongly dependent on task characteristics and task sequence.

## 1. Introduction

Previous knowledge of the time a component's failure will occur allows for countermeasures and reduces the risk to a system's safety and security [1]. In case of high-volume components such as lithium-ion batteries this obviously has a high economic and ecologic relevancy.

However, taking those lithium-ion batteries as an example, their fault mechanisms and usage behaviors are so complex, that traditional, model-driven approaches do not suffice [2]. Data-driven approaches not relying on explicit knowledge of the specific wear mechanisms offer help but are themselves challenged by acquiring sufficient training data and staying up to date with a problem space growing with every new usage pattern or operating condition [3].

Mitigation to this problem could be provided by knowledge transfer between continuously learning, generalizing algorithms, bridging over gaps between different smaller datasets and accumulating knowledge over time [4]. Such a transfer could be realized using regularization-based continual learning approaches, which allow sequential learning of similar tasks without overwriting previously acquired knowledge [5, 6].

*Objective*: In this article, the feasibility of different regularization approaches towards solving sequential learning problems is analyzed using a time series benchmark dataset on lithium-ion battery wear.

*Structure*: Chapter 2 presents related work on the topics of fault prediction for lithium-ion batteries and regularization-based continual learning. From there, a methodology is derived in chapter 3. A dataset is introduced and experiments conducted on that dataset are described together with their results in chapter 4. Concludingly, chapter 5 summarizes the main findings and presents an outlook.

**2. Related work**

*2.1. Fault Prediction for Lithium-Ion Batteries*

Lithium-ion batteries are becoming truly ubiquitous, powering all sorts of mobile, rechargeable electrical appliances – from small ear phones to cars [7] and possibly even aircraft [8]. Naturally, such batteries are subject to wear, which is caused by a number of different processes while cycling or even resting, with many different parameters having an impact [2]. Due to the large volume of those batteries, understanding and predicting their wear is an extremely important task, both economically and ecologically. However, the complex system of intertwined wear processes is not understood yet [2].

Data-driven fault prediction aims at solving this problem without understanding the underlying physics, predicting the remaining useful lifetime (RUL), i.e. the time to failure, of an entity purely based upon training data. This regression task must be differentiated from a fault diagnosis' categorization as it is oblivious to the cause of failure [1]. There are numerous examples of deep-learning-based fault prediction one the problem of lithium-ion battery wear:

In [9], a long short-term memory (LSTM) approach was enhanced by resilient backpropagation, dropouts and a Monte Carlo simulation which provides prediction confidence information. Self-measured capacity data of six 18650 lithium-ion batteries cycled under different temperatures and current-rates was used to carry out the multi-step ahead prediction of the same value which was used to calculate the RUL. Experiments included comparisons of online and (semi-)offline training capabilities with the proposed approach showing good results on both.

In [10], a fully connected auto-encoder was used on a large battery degradation dataset by NASA [11]. From each charging or discharging, 21 human-engineered features, e.g. terminal voltage or output current, are selected as an input. However, despite data from only three batteries examined under identical conditions being used, the results were still mediocre.

In [12], a thorough comparison of differently parametrized feed-forward neural networks, convolutional neural networks (CNN) and an LSTM network was carried out on the aforementioned NASA dataset [11] with the goal of predicting capacities. As inputs, sub-sampled, normalized voltage (V), current and surface temperature data of four batteries examined under identical conditions were used. A further examination of utilizing single (only V) as opposed to the aforementioned multi-channel input information was carried out. The multi-channel approach resulted in better performances with LSTM achieving the overall best result.

Despite these promising results, as [3] points out, deep learning based approaches' performance heavily relies on the training process, which in turn relies on the availability of data

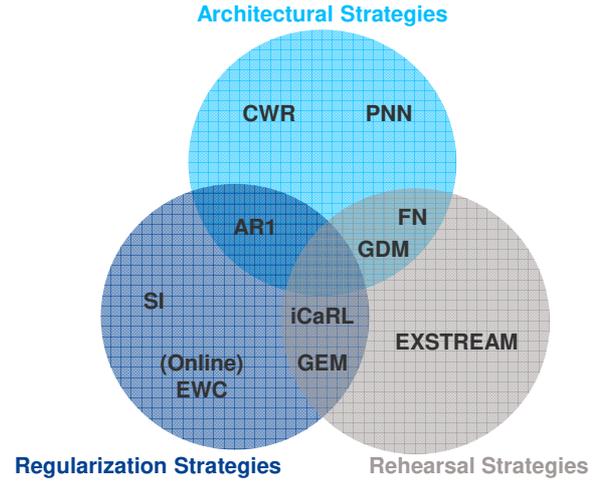

Fig. 1. Venn diagram of some of the most popular continual learning strategies based upon [17] (CWR: CopyWeights with Re-Init; PNN: Progressive Neural Networks; FN: FearNet; GDM: Grwoing Dual-Memory; iCaRL: Incremental Classifier and Representation Learning; ExStream: Exemplar Streaming; GEM: Gradient Episodic Memory)

and its representativeness of the task. The fact that e.g. [12] only uses batteries of one of nine experimental groups bears witness to the practical relevance of this challenge. To solve it, self-improving models via online data are suggested [1, 5].

*2.2. Regularization-based continual learning*

In machine learning, the transfer of knowledge and skills from one or more source tasks to a target task in order to train a deep learning algorithm capable of solving both, source and target tasks, is referred to as 'continual learning' [4, 13]. In the field of fault prediction, this can facilitate learning across several smaller, less homogenous datasets [5], mitigating two key problems hindering a more widespread utilization of machine learning [4]:

- Because of the high diversity of possible operating conditions and their near-to unforseeability for device manufacturers combined with high standards of privacy shielding user data, acquiring datasets sufficiently large and diverse for successful training is difficult [14, 15].
- Because of a theoretically ever-expanding database of new operating conditions and usage behaviors, even if a sufficient dataset could be acquired, it would only provide short-term representations of the problem space necessitating continuous data collection and retraining of algorithms [16].

There are three categories of continual learning strategies commonly distinguished: architectural, rehearsal and regularization strategies [17] (see Fig. 1). For mitigating the two above-described problems, one appears to be suited best: Whereas *rehearsal strategies* still rely on sharing of at least some data and *architectural strategies* strive only on more loosely related tasks, *regularization strategies* use altered loss functions in order to solve more closely related tasks, promising to allow generalization over different e.g. battery usage scenarios without the need for an exchange of potentially confidential raw data.

Modelled after the process of synaptic consolidation in a mammalian brain, *regularization strategies* slow down the change of certain weights depending on their importance on previously learned tasks, thereby selectively reducing the network's plasticity.

Three specific implementations of regularization strategies are commonly included in comparative analyses [17–19]:

*Elastic weight consolidation* (EWC) is based on the idea [20] that more than one set of weights $\theta$ represents a possible solution $\theta_A$ of a task A, so that a solution $\theta_{AB}$ can be found that solves both tasks A and B [21]. This is achieved by adding a penalty to the loss function (see Eq. 1): $L_C(\theta_{ABC})$ is the (conventional) loss for task C on a set of weights $\theta_{ABC}$ capable to solve all tasks A, B and C, $\lambda$ defines the importance of old tasks compared to the new one, $F$ is the diagonal of the Fisher information matrix and $i$ labels each individual parameter.

$$L(\theta_{ABC}) = L_C(\theta_{ABC}) + \lambda \cdot \sum_i [F_{A,i}(\theta_{ABC,i} - \theta_{A,i}^*)^2 + F_{B,i}(\theta_{ABC,i} - \theta_{B,i}^*)^2] \quad (1)$$

*Online EWC* expands on this idea, but shifts attention from older to newer tasks by not relying on Fisher information matrices for every tasks but on only one for all tasks combined [22]. This reduces Eq. 1 to Eq. 2:

$$L(\theta_{ABC}) = L_C(\theta_{ABC}) + \lambda \cdot \sum_i F_{AB,i}(\theta_{ABC,i} - \theta_{AB,i}^*)^2 \quad (2)$$

*Synaptic intelligence* (SI) relies on a similar idea, but uses an importance measure $\omega$ that is calculated directly during the stochastic gradient descent as opposed to the Fisher information matrix which needs to be calculated separately [23], leading to Eq. 3:

$$L(\theta_{ABC}) = L_C(\theta_{ABC}) + \lambda \cdot \sum_i \omega_{AB,i}(\theta_{ABC,i} - \theta_{AB,i}^*)^2 \quad (3)$$

Although the aforementioned regularization approaches displayed good results on different general evaluation datasets [18, 19], there are to the authors' knowledge no publications on their performance in the industrial domain – with two exceptions being EWC for fault prediction [5] and all three approaches for anomaly detection [6]. This article therefore aims to provide a first comparative analysis of those approaches on a fault prediction task.

## 3. Methodology

Data preprocessing has a substantial influence on any deep learning algorithm's performance [24]. In order to circumvent problems arising from time series of different lengths (e.g. during charging and discharging) or from sub-sampling away important information, similar to [10] an extraction of 21 human-engineered, battery-specific features was carried out (10 for charging, 11 for discharging).

Using these features and building upon the good performance of recurrent neural networks presented in chapter 2.1 combined with the desire to create a simple algorithm

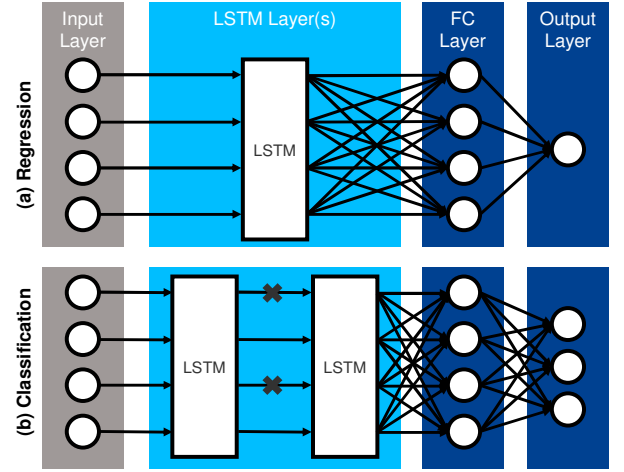

Fig. 2. Architecture of the base algorithms' deep neural networks for the non-regularized regression task (a) and the regularized classification task (b)

requiring neither vast computing resources nor extensive optimization, a multilayer LSTM-approach was chosen as base algorithm. An ensuing hyperparameter optimization yielded the structure depicted in Fig. 2 (Regression) and parameters listed in Table 1 (Regression).

Upon this base algorithm, the different regularization approaches were implemented. Because RUL prediction is a regression problem and regularization requires prediction likelihoods only available for classification problems [5], the base algorithm needs to be altered. The translation of the continuous RUL scale into categories was carried out as listed in Table 2.

For regularization-based approaches, initial tests resulted in online EWC performing best. Therefore, a hyperparameter optimization for the base algorithm's parameters was carried out based upon this approach, resulting in the structure depicted in Fig. 2 (Classification). Then, another hyperparameter optimization for the different approaches' parameters was conducted. The results are listed in Table 1 (Classification).

Table 1: Hyperparameters used for the different regularization-based continual learning algorithms

| Parameter | Value (Regression) | Value (Classification) |
|---|---|---|
| Batch size | 5 | 5 |
| Number of hidden layers | 1 | 2 |
| Number of nodes per hidden layer | 150 | 100 |
| Drop-out probability | 0.0 | 0.1 |
| $\lambda_{EWC}$ | - | 500,000 |
| $\gamma_{EWC}$ | - | 2 |
| $c_{SI}$ | - | 200 |

Table 2: Definition of classes by remaining useful life (RUL) values and corresponding state-of-health (SoH) label

| Class No. | SoH-Label | Condition |
|---|---|---|
| 1 | "safe" | RUL > 60 |
| 2 | "okay" | 30 < RUL ≤ 60 |
| 3 | "at risk" | RUL ≤ 30 |

## 4. Experiments

In this chapter, an open access dataset collected from a battery test rig is introduced. Using this dataset, different experiments are conducted, first testing the non-regularized approach on a continuous (regression) RUL scale before examining the regularization approaches in order to classify samples by state-of-health (SoH), an incremental task learning scenario [18, 19, 25].

All experiments were conducted on a computer featuring an AMD Ryzen Threadripper 2920X CPU and a NVIDIA GeForce RTX 2080 8 GB GPU running Ubuntu 20.04. The learning framework used was PyTorch 1.6 under Python 3.6.

### 4.1. Experimental dataset

The experiments were conducted using a subset of an open access lithium-ion battery degradation dataset. The complete dataset consists of data from 34 batteries organized in 9 experimental groups of 3 to 4 batteries each. Within one experimental group, all batteries were subject to the same parameters and conditions. The batteries were cycled, i.e. charged and discharged, under controlled environmental conditions until their remaining capacity dropped below an experimental-group-specific threshold value. During charging and discharging, data regarding e.g. battery voltage, current or surface temperature was collected (see Fig. 3). Depending on external factors, such as ambient temperature or charging mode, which were also monitored, the batteries experienced different levels of wear.

Because of missing end-of-life-conditions (which are necessary for labeling the data) and missing or obviously incorrect measurements, only experimental groups 1 (batteries 5, 6, 7 and 18), 4 (batteries 33, 34 and 36), 7 (batteries 45 to 48) and 9 (batteries 53 to 56) were used.

In addition to the previously mentioned challenge of learning to predict battery wear across different environmental and usage conditions, this dataset consist of only a few hundred cycles per condition – an amount of data sometimes considered to be too little for deep learning [26].

### 4.2. Conventional deep-learning-based fault prediction (Regression)

On the continuous regression RUL scale, the non-regularized approach was tested in two ways: Firstly, batteries 5 and 6 were used for training and battery 7 for testing. Secondly, all three datasets were mixed, 80% were used for training and 20% for testing. In both cases, training was conducted over 50 epochs.

Table 3 lists the results: In the first case, the proposed approach easily outperforms the published results by [10]. With a root mean square error (RMSE) of just 4.81 %, the performance is already very good. However, mixing the different battery datasets as in case 2, the RMSE further decreases to just 2.15 %. To put it into more practical terms, the mean deviation is just 9,1 usage cycles in case 1 or 2.5 usage cycles in case 2.

Fig. 4 shows the resulting graphs: In both cases, the deviation stays inside a narrow band around the actual values with the prediction accuracy in case 2 being clearly better.

Apparently, even in this scenario of testing identical batteries under identical circumstances the differences between the individual batteries are still considerable. This highlights the need for robust knowledge transfer, because retraining a prediction algorithm from scratch every time a new dataset becomes available is clearly uneconomical.

### 4.3. Regularization-based continual learning fault prediction (Classification)

On the classification task of predicting the current SoH, the regularized approaches described in section 3 were tested: Each experimental group was considered to be a semi-independent task. Randomly drawn 80% of the data from any experimental group were used for training, 20% were used for testing. The algorithms were trained sequentially for 100 epochs each on experimental groups 1, 7, 4 and 9 (in this order). For baseline comparison, an algorithm without any regularization enhancement was used. After training on one task, the algorithm was validated on all tasks before switching to the next task. All stated values are mean values collected over five runs of the experiment.

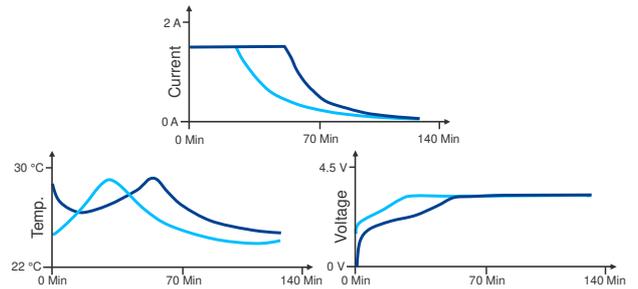

Fig. 3. Comparison of data from a fresh (dark blue) and an aged (light blue) lithium-ion battery cell

Table 3: Overall RUL prediction accuracy using only experimental group 1 (Case 1: Batteries 5, 6 for training and battery 7 for testing; Case 2: 80% of batteries 5, 6, 7 for training, 20% of batteries 5, 6, 7 for testing)

| Approach | RMSE (Case 1) | RMSE (Case 2) |
|---|---|---|
| FC-AE [10] | 11.8 % | - |
| Proposed approach: LSTM | 4.81 % | 2.15 % |

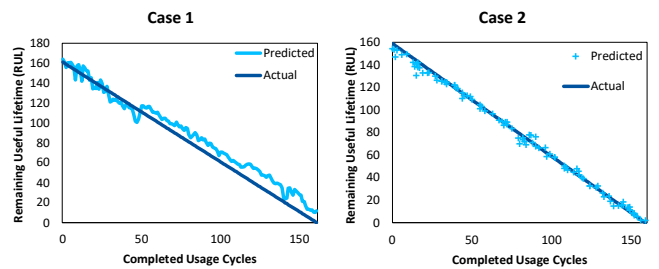

Fig. 4. RUL prediction test results for case 1 (left) and case 2 (right)

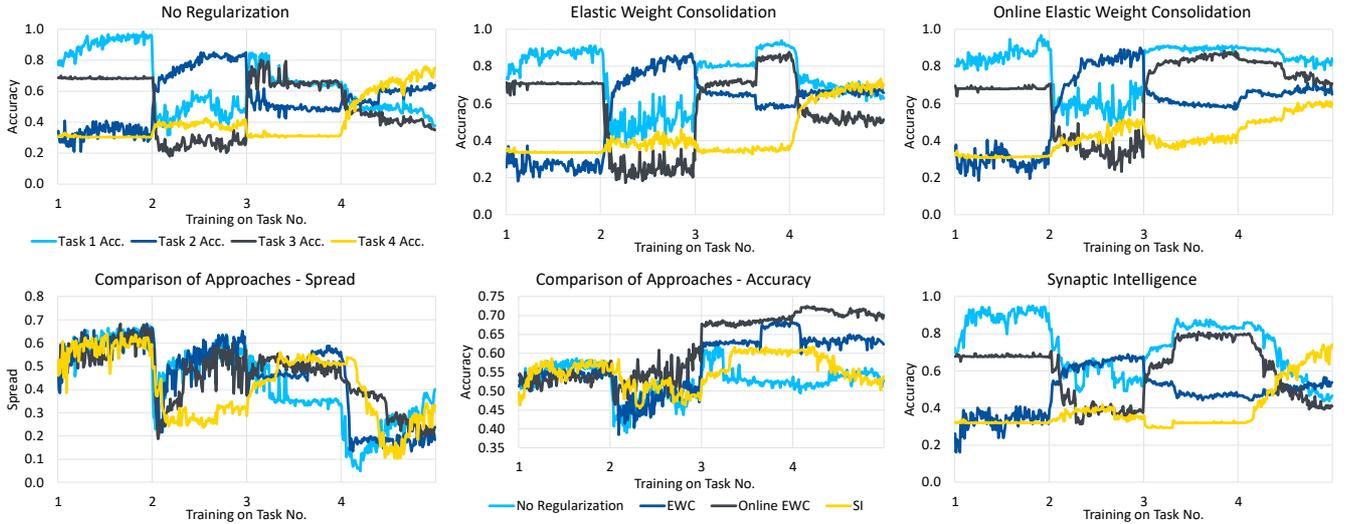

Fig. 5. Results on a transfer learning scenario involving datasets from four different experimental groups subject to four different usage scenarios using no regularization, EWC, online EWC or SI approaches (clockwise, starting top left; legend only in first diagram) as well as a comparison of mean accuracies (bottom middle, with legend) and maximum spread (bottom left, for legend see mean accuracy figure) over five runs

*Without regularization* (see Fig. 5, top left diagram), the maximum accuracy on the task being trained on is usually the highest. It ranges between 0.97 for task 1 and 0.75 for task 4. Surprisingly, on task 3 it achieves its best results in the beginning of its training, while on the others it improves throughout the entire training. The rapid decline of one task's accuracy after training switches to the next task, the so-called catastrophic forgetting [27], is easy to distinguish (e.g. when switching from training on task 1 to training on task 2). Overall, even after training on all tasks, the algorithm is clearly not capable of solving all of them sufficiently well, underlining the need for a different approach.

When *EWC* is used (see Fig. 5, top middle diagram), the maximum accuracy on the task being trained on is slightly declining from 0.91 for task 1 to 0.73 for task 4. Surprisingly, while training on task 3 the algorithm achieves its best accuracy on task 1. After training switches from on task to the next, the prior tasks' accuracies decline – although considerably less than without any regularization.

When *Online EWC* is used (see Fig. 5, top right diagram), the maximum accuracy on the task being trained on is slightly declining from 0.95 for task 1 to 0.61 for task 4. Surprisingly, while training on task 3 and 4 the algorithm achieves its best accuracy on task 1. After training switches from on task to the next, the prior task's accuracy declines, but stays somewhat constant following subsequent switches. This causes the final accuracies of all tasks except the last being trained on to be significantly higher than without any regularization or using EWC.

When *SI* is used (see Fig. 5, bottom right diagram), the maximum accuracy on the task being trained on ranges from 0.95 (task 1) to 0.69 (task 2). Surprisingly, the algorithm performs worst while training on task 2 (mean accuracy over all tasks being 0.51) and achieves its best accuracy while training on task 3 on task 1.

Comparing the *mean accuracies over all four tasks* (see Fig. 5, bottom middle diagram), all approaches start out similarly during training on task 1. During training on task 2, Online EWC starts to show better results. After a significant, initial increase, mean accuracies stay relatively constant during training on tasks 3 and 4 with Online EWC slightly improving and SI slightly declining over time.

Comparing the *maximum spread over all four tasks* (see Fig. 5, bottom left diagram), all approaches start out similarly with between 0.5 to 0.7 spread between best and worst tasks' accuracies during training on task 1. During training on task 2, only SI has a considerably lower spread (approx. 0.3) than the other approaches (approx. 0.5) – unfortunately caused by its low accuracy on all tasks. During training on task 3, the no-regularization approach has a considerably lower spread (approx. 0.35), which, again, is caused by its low accuracy on all tasks. During training on task 4, all algorithms' spreads significantly decline initially (to approx. 0.2), with SI's and no-regularization's rebounding towards the end of training.

Table 4 focusses on the different tasks' *mean accuracies on each task over the last 20 epochs* of training on task 4: Although it has the overall highest accuracy on task 4 (its best task at 0.72), no-regularization expectably fares worst regarding the lowest (0.38 for task 3) and the mean accuracies (0.54). SI performs slightly worse on maximum and mean accuracy, although its worst accuracy is marginally better. With the same accuracy on task 4 (0.7), EWC has much better lowest (0.51 for task 3) and mean (0.63) accuracies. Concludingly, Online EWC delivered the highest values for best, mean and worst accuracy (0.82 at task 1, 0.7 and 0.6 at task 4).

Table 4: Mean accuracy over last 20 epochs on four tasks after training on four tasks

| Approach | Best task | Mean | Worst task |
|---|---|---|---|
| No Regularization | 0.72 | 0.54 | 0.38 |
| Elastic Weight Consolidation | 0.70 | 0.63 | 0.51 |
| Online Elastic Weight Consolidation | **0.82** | **0.70** | **0.60** |
| Synaptic Intelligence | 0.70 | 0.52 | 0.41 |

Online EWC clearly performs best in learning to solve all four tasks. However, the decreasing accuracy on the task being currently trained on raises the question of its performance when more tasks are added to the sequence.

Furthermore, there seems to be a strong correlation between the tasks 1 and 3 (i.e. experimental groups 1 and 4) as training of one benefitted the other on all methods. Contrastingly, training on task 2 (i.e. experimental group 7) resulted in pronounced catastrophic forgetting for tasks 1 and 3 – even for regularization methods more resistant to forgetting – with only a marginal improvement for task 4.

## 5. Conclusion

In this paper, the feasibility of different regularization approaches towards solving sequential learning problems in industrial use cases was examined. An open-access time series benchmark dataset on lithium-ion battery wear was used to evaluate the algorithms, because, here, data-driven fault prediction promises an especially wide-spread applicability.

Using an LSTM approach, a new high mark on the non-sequential regression task of predicting remaining useful lifetime (RUL) for lithium-ion batteries could be established.

For the sequential learning problem, the continuous RUL scale had to be converted into a distinct state-of-health classes first. Then, three different regularization approaches could be evaluated on the sequential learning task. Our main findings are:

- Regularization approaches improve our base algorithm's performance compared to no regularization.
- The online elastic weight consolidation approach outperforms the elastic weight consolidation and synaptic intelligence approaches.
- The tasks themselves have a big influence on the learning performance. It remains unclear, whether this only depends on their relative similarity to each other or also on their individual position in the learning sequence.

Future research should carry out hyperparameter optimizations for all approaches, possibly even for different task sequence lengths. Furthermore, the possibility of one-shot learning using regularization methods should be examined. Additionally, experiments involving other industrial datasets could be of interest.

## References

[1] Xu G et al. Data-Driven Fault Diagnostics and Prognostics for Predictive Maintenance: A Brief Overview. 2019 IEEE 15th International Conference on Automation Science and Engineering (CASE), Vancouver; 2019; 103–8.

[2] Kabir M, Demirocak D. Degradation mechanisms in Li-ion batteries: a state-of-the-art review. Int J Energy Res 2017; 14:1963–86.

[3] Li Y et al. Data-driven health estimation and lifetime prediction of lithium-ion batteries: A review. Renewable and Sustainable Energy Reviews 2019; p. 109254.

[4] Maschler B, Weyrich M. Deep Transfer Learning for Industrial Automation. Industrial Electronics Magazine 2021; 2:(in print).

[5] Maschler B, Vietz H, Jazdi N, Weyrich M. Continual Learning of Fault Prediction for Turbofan Engines using Deep Learning with Elastic Weight Consolidation. 2020 25th IEEE International Conference on Emerging Technologies and Factory Automation (ETFA), Vienna; 2020; 959–66.

[6] Maschler B, Pham T, Weyrich M. Regularization-based Continual Learning for Anomaly Detection in Discrete Manufacturing. Preprint: 2021.

[7] Hu X, Zou C, Zhang C, Li Y. Technological Developments in Batteries: A Survey of Principal Roles, Types, and Management Needs. IEEE Power and Energy Mag. 2017; 5:20–31.

[8] Chin J et al. Battery Evaluation Profiles for X-57 and Future Urban Electric Aircraft. AIAA Propulsion and Energy 2020 Forum, Virtual Event; 2020; 1–13.

[9] Zhang Y, Xiong R, He H, Pecht M. Long Short-Term Memory Recurrent Neural Network for Remaining Useful Life Prediction of Lithium-Ion Batteries. IEEE Trans. Veh. Technol. 2018; 7:5695–705.

[10] Ren L et al. Remaining Useful Life Prediction for Lithium-Ion Battery: A Deep Learning Approach. IEEE Access 2018; 50587–98.

[11] Saha B, Goebel K. Battery Data Set: NASA AMES Prognostics Data Repository; 2007.

[12] Choi Y, Ryu S, Park K, Kim H. Machine Learning-Based Lithium-Ion Battery Capacity Estimation Exploiting Multi-Channel Charging Profiles. IEEE Access 2019; 75143–52.

[13] Parisi G et al. Continual lifelong learning with neural networks: A review. Neural Networks 2019; 113:54–71.

[14] Tercan H et al. Transfer-Learning: Bridging the Gap between Real and Simulation Data for Machine Learning in Injection Molding. Procedia CIRP 2018; 185–90.

[15] Maschler B, Jazdi N, Weyrich M. Maschinelles Lernen für intelligente Automatisierungssysteme mit dezentraler Datenhaltung am Anwendungsfall Predictive Maintenance. VDI Reports 2019; 2351:739–51.

[16] Tercan H, Guajardo A, Meisen T. Industrial Transfer Learning: Boosting Machine Learning in Production. Proceedings of the 2019 IEEE 17th International Conference on Industrial Informatics (INDIN), Helsinki; 2019; 274–9.

[17] Maltoni D, Lomonaco V. Continuous learning in single-incremental-task scenarios. Neural Networks 2019; 116:56–73.

[18] Hsu Y-C, Liu Y-C, Ramasamy A, Kira Z. Re-evaluating Continual Learning Scenarios: A Categorization and Case for Strong Baselines. 32nd Conference on Neural Information Processing Systems (NeurIPS) Continual Learning Workshop 2018.

[19] van de Ven G, Tolias A. Three scenarios for continual learning. 32nd Conference on Neural Information Processing Systems (NeurIPS) Continual Learning Workshop 2018.

[20] Hinton G, Vinyals O, Dean J. Distilling the Knowledge in a Neural Network. 28th Conference on Neural Information Processing Systems (NeurIPS) Continual Learning Workshop 2014.

[21] Kirkpatrick J et al. Overcoming catastrophic forgetting in neural networks. Proceedings of the National Academy of Sciences of the United States of America 2017; 13:3521–6.

[22] Schwarz J et al. Progress & Compress: A scalable framework for continual learning. Proceedings of Machine Learning Research 2018; 80:4528–37.

[23] Zenke F, Poole B, Ganguli S. Continual Learning Through Synaptic Intelligence. Proceedings of Machine Learning Research 2017; 70:3987–95.

[24] Maschler B, Ganssloser S, Hablizel A, Weyrich M. Deep learning based soft sensors for industrial machinery. Procedia CIRP 2021; 99:662–7.

[25] Maschler B, Kamm S, Weyrich M. Deep Industrial Transfer Learning at Runtime for Image Recognition. at - Automatisierungstechnik 2021; 3:211-220.

[26] Eker O, Camci F, Jennions I. Major Challenges in Prognostics: Study on Benchmarking Prognostics Datasets. 2012 1st European Conference of the Prognostics and Health Management Society (PHM-E), Dresden; 2012; 148–55.

[27] French R. Catastrophic forgetting in connectionist networks. Trends in Cognitive Sciences 1999; 4:128–35.